\documentclass[sigconf]{acmart}

\usepackage{adjustbox}
\usepackage{multirow}
\usepackage{balance}

\AtBeginDocument{%
  \providecommand\BibTeX{{%
    \normalfont B\kern-0.5em{\scshape i\kern-0.25em b}\kern-0.8em\TeX}}}

\copyrightyear{2024}
\acmYear{2024}
\setcopyright{acmlicensed}\acmConference[MAD '24]{3rd ACM International Workshop on Multimedia AI against Disinformation}{June 10, 2024}{Phuket, Thailand}
\acmBooktitle{3rd ACM International Workshop on Multimedia AI against Disinformation (MAD '24), June 10, 2024, Phuket, Thailand}
\acmDOI{10.1145/3643491.3660277}
\acmISBN{979-8-4007-0552-6/24/06}

\begin{document}

\title{SIDBench: A Python Framework for Reliably Assessing Synthetic Image Detection Methods}

\author{Manos Schinas}
\email{manosetro@iti.gr}
\orcid{0009-0007-2132-3430}
\affiliation{%
  \institution{Information Technologies Institute @ CERTH}
  \streetaddress{6th km Harilaou - Thermis, 57001}
  \city{Thessaloniki}
  \country{Greece}
  \postcode{43017-6221}
}

\author{Symeon papadopoulos}
\email{papadop@iti.gr}
\orcid{0000-0002-5441-7341}
\affiliation{%
  \institution{Information Technologies Institute @ CERTH}
  \streetaddress{6th km Harilaou - Thermis, 57001}
  \city{Thessaloniki}
  \country{Greece}
}


\begin{abstract}
The generative AI technology offers an increasing variety of tools for generating entirely synthetic images that are increasingly indistinguishable from real ones. 
Unlike methods that alter portions of an image, the creation of completely synthetic images presents a unique challenge and several Synthetic Image Detection (SID) methods have recently appeared to tackle it. Yet, there is often a large gap between experimental results on benchmark datasets and the performance of methods in the wild. To better address the evaluation needs of SID and help close this gap, this paper introduces a benchmarking framework that integrates several state-of-the-art SID models. 
Our selection of integrated models was based on the utilization of varied input features, 
and different network architectures, aiming to encompass a broad spectrum of techniques. The framework leverages recent datasets with a diverse set of generative models, high level of photo-realism and resolution, reflecting the rapid improvements in image synthesis technology. 
Additionally, the framework enables the study of how image transformations, common in assets shared online, such as JPEG compression, affect detection performance. SIDBench is available on \href{https://github.com/mever-team/sidbench}{github.com/mever-team/sidbench} and is designed in a modular manner to enable easy inclusion of new datasets and SID models.

\end{abstract}





\maketitle

\section{Introduction}

In recent years, a growing number of image generation models have emerged that enable low-effort generation of synthetic images with an unprecedented level of realism. Such images are increasingly indistinguishable from authentic ones, introducing significant challenges to the integrity of digital content. This is particularly evident in the realms of political discourse and propaganda, where the ability to fabricate convincing synthetic images can manipulate public perception on important issues. Examples from recent events, such as the Ukraine war or the Gaza conflict, illustrate the power of synthetic images in shaping convincing narratives. In these contexts, generative models have been employed to create images that either exaggerate the situation, depict events that never occurred, or manipulate the portrayal of political figures.

Image manipulation has evolved with the advent of digital editing software and tools. Unlike conventional digital manipulation, which involves manually editing parts of an original image, the creation of fully synthetic images introduces a unique challenge. These images lack the ``traditional'' forensic traces that could reveal a manipulated photograph. As such, they require a different set of techniques for detection, pushing the boundaries of image forensics technology. Therefore, in this work, we focus on detecting synthetic images and omit discussions on forgery detection and localization, areas already covered extensively in the literature \cite{mehrjardi2023survey,zanardelli2023image,zampoglou2017large}. 

Generative models have greatly improved synthetic image creation, with developments in Variational Autoencoders (VAEs), Generative Adversarial Networks (GANs) \cite{goodfellow2014generative} and Diffusion Models (DMs) \cite{sohl2015deep,yang2023diffusion}, being the most notable and impressive. Diffusion-based models have emerged as a new standard in the image generation domain; 
recent text-to-image diffusion models, such as Dall$\cdot$E \cite{ramesh2022hierarchical,ramesh2021zero}, Glide \cite{pmlr-v162-nichol22a}, Stable Diffusion \cite{Rombach_2022_CVPR}, and Imagen \cite{saharia2022photorealistic} have reached impressive results in terms of image quality and realism. Additionally, commercial tools like Midjourney\footnote{\url{https://www.midjourney.com}} and Adobe Firefly\footnote{ \url{https://www.adobe.com/gr_en/products/firefly.html}}  enable users to create fake images with a high degree of photo-realism without requiring technical skills.

These developments attract growing interest in methods for Synthetic Image Detection (SID). However, due to large variations among generative models, training universal detection models has proven to be an extremely hard problem. Generalization to unseen generative models remains a challenge. Another significant issue is evaluating all these proposed models in the wild. Typically, synthetic images shared online are of high quality in contrast to images used for evaluation in the lab. Furthermore, understanding which methods work well under specific circumstances is also challenging. For instance, some models are sensitive to image transformations, leading to high rates of false positives or negatives. 

To address the growing need of assessing SID methods in a realistic manner, we propose the SIDBench framework. 
This is designed with flexibility in mind, allowing for easy extension with additional detection models and datasets. SIDBench 
provides a comprehensive performance comparison among several state-of-the-art models, and reveals the nuances of each model, indicating under which conditions they perform best or fall short. Its modularity and ease of support for incorporating new models and datasets ensures it can remain relevant in the future, facilitating continuous assessment as new generative and detection models emerge. SIDBench is available on \href{https://github.com/mever-team/sidbench}{GitHub} to help standardise future research in this area.

\section{Related Work}

\subsection{Synthetic Image Generation} 

The landscape of image generation has seen significant advances, primarily driven by three approaches: Variational Autoencoders (VAEs), Generative Adversarial Networks (GANs) \cite{10.1145/3422622}, and Diffusion Models (DMs) \cite{pmlr-v37-sohl-dickstein15}. These developments have made it possible to create highly realistic synthetic images. Until recently, synthetic images were mainly generated using GANs, which can be categorized into conditional and unconditional frameworks. Conditional GANs such as CycleGAN \cite{Zhu_2017_ICCV}, StarGAN \cite{Choi_2018_CVPR}, and GauGAN \cite{Park_2019_CVPR}, generate images based on given conditions or labels, offering more control over the output, while unconditional GANs, such as ProGAN \cite{karras2018progressive}, StyleGAN \cite{Karras_2019_CVPR}, and BigGAN \cite{brock2018large}, generate images from random noise. While GANs have deeply influenced the domain of image generation, they have recently been superseded by DMs. 
DMs for image generation can be categorized primarily into Denoising Diffusion Probabilistic Models (DDPMs) \cite{NEURIPS2020_4c5bcfec,pmlr-v139-nichol21a}, which straightforwardly reverse the diffusion process, and Latent Diffusion Models (LDMs) \cite{Rombach_2022_CVPR}, optimizing the process within a compressed latent space for efficiency. A notable unconditional approach is the Ablated Diffusion Model (ADM) \cite{NEURIPS2021_49ad23d1} that has recently exceeded the image generation capabilities of GANs. More recently, conditional diffusion (text-to-image) models gained popularity. Models in this category take a text prompt and random noise as input and then denoise the image under the guidance of the prompt so that the generated image matches the description. Among those, Guided Diffusion \cite{NEURIPS2021_49ad23d1} and Glide \cite{pmlr-v162-nichol22a} are two very popular methods. 

\subsection{Synthetic Image Detection} 

SID focuses on identifying whether an image is authentic or artificially created (e.g., produced by generative models. This task is often approached as a binary classification problem, where a pre-trained backbone model such as ResNet50 is further fine-tuned on the real/fake task. These classifiers, such as the work of \cite{Wang_2020_CVPR}, typically perform well when the fake images for testing and training are from the same generative distribution. However, their effectiveness decreases on unseen generative models not encountered during training. The works of \cite{Wang_2020_CVPR} and \cite{corvi2023detection} suggested that appropriate data augmentation techniques can significantly improve generalization. Other works use fingerprints extracted from images under the assumption that fingerprint patterns differ between synthetic and real images. The work of \cite{pmlr-v119-frank20a} uses patterns extracted from the frequency domain of the images, as the up-sampling operation in GAN-based image generation introduces detectable frequency artifacts. Other fingerprint patterns for effective generalization include noise patterns in the frequency domain \cite{10.1007/978-3-031-19781-9_6} and texture patterns \cite{Liu_2020_CVPR,zhong2024patchcraft}. A recent study \cite{Wang_2023_ICCV} observed that synthetic images from DMs could be more accurately reconstructed by other DMs, utilizing reconstruction errors as a unique fingerprint. Additionally, a recent approach \cite{Ojha_2023_CVPR} explored the use of detectors based on frozen CLIP image encoders, which demonstrate superior generalization across unseen model families compared to ResNet-based detectors.

\section{SIDBench Framework}

SIDBench is a Python project that incorporates several open-source detection models developed with PyTorch. Due to the differing implementation details and structural variations across these projects, SIDBench introduces a common wrapper interface, facilitating the addition of new models. Consequently, while integrating new models with SIDBench requires some initial effort, it significantly simplifies the subsequent evaluation process. To study the performance of the incorporated models, SIDBench utilizes well-known datasets frequently cited in the related literature, with the appropriate data loaders. Given that the task is binary classification, implementing a PyTorch DataLoader that adheres to the structure of a new dataset is straightforward. This streamlined approach enables users to efficiently assess models' performance across diverse datasets, thereby providing a robust evaluation of their effectiveness.

\subsection{Evaluation Data}

For evaluation, we considered a variety of generative models. First, we incorporated the generative models used in \cite{Wang_2020_CVPR}. The training set of this dataset consists of ProGAN images, and has been used for the training of all GAN-based detectors in our benchmark. The testing dataset includes ProGAN \cite{karras2018progressive}, StyleGAN \cite{Karras_2019_CVPR}, BigGAN \cite{brock2018large}, CycleGAN \cite{Zhu_2017_ICCV}, StarGAN \cite{Choi_2018_CVPR}, GauGAN \cite{Park_2019_CVPR}, CRN \cite{Chen_2017_ICCV}, IMLE \cite{Li_2019_ICCV}, SAN \cite{Dai_2019_CVPR}, SITD \cite{Chen_2018_CVPR}, and DeepFakes \cite{Rossler_2019_ICCV}. In addition we used the dataset from \cite{Ojha_2023_CVPR} that includes recent text-to-image generation models for the generation of 256×256 resolution images. The Latent diffusion model (LDM) \cite{Rombach_2022_CVPR} and Glide \cite{pmlr-v162-nichol22a} are DM variants, and DALL$\cdot$E is an auto-regressive model. Following the setting in \cite{Ojha_2023_CVPR}, three variants of LDM were used for evaluation: (i) LDM with 200 denoising steps, (ii) LDM with 200 steps with classifier-free diffusion guidance (CFG), and (iii) LDM with 100 steps. Similarly, three variants of the Glide model have been used based on the number of steps in the two stages of noise refinement: (i) 100 steps in the first stage followed by 27 steps in the second stage (100-27) \footnote{This is the standard practice were 100 steps are used to get a low resolution image of 64×64 and then 27 steps are used to up-sample images to 256×256.}, (ii) 50-27, and (iii) 100-10. Images of the LAION dataset \cite{schuhmann2021laion} are used for the real class. Table \ref{tab:datasets} summarizes these two sources. 

\begin{table}[]
\caption{Test datasets from \cite{Wang_2020_CVPR} and \cite{Ojha_2023_CVPR}.}
\label{tab:datasets}
\scalebox{0.92} {
\begin{tabular}{lccr}
\toprule
Family & Method & Source & \#Images \\
\midrule
\multirow{3}{*}{\begin{tabular}[c]{@{}l@{}}Unconditional\\ GAN\end{tabular}} & ProGAN & LSUN & 8.0k \\
 & StyleGAN & LSUN & 12.0k \\
 & StyleGAN2 & LSUN & 16.0k \\
 & BigGAN & ImageNet & 4.0k  \\
 \midrule
\multirow{3}{*}{\begin{tabular}[c]{@{}l@{}}Conditional\\ GAN\end{tabular}} & CycleGAN & - & 2.6k  \\
 & StarGAN & CelebA & 4.0k \\
 & GauGAN & COCO & 10.0k \\
 \midrule
\begin{tabular}[c]{@{}l@{}}Perceptual\\ loss\end{tabular} & CRN & GTA & 12.8k \\
 & IMLE & GTA & 12.8k  \\
\begin{tabular}[c]{@{}l@{}}Low-level\\ vision\end{tabular} & SITD & Raw camera & 0.36k \\
 & IMLE & Standard SR & 0.44k  \\
\midrule
Deepfake & FaceForensics++ & Videos of faces & 5.4k \\
\midrule
\multirow{2}{*}{\begin{tabular}[c]{@{}l@{}}Text-to-image \\ Diffusion\end{tabular}} & Latent Diffusion & LAION & 3.0k \\
 & GLide & LAION &  3.0k \\
\begin{tabular}[c]{@{}l@{}}Guided\\ Diffusion \end{tabular} & Guided \cite{NEURIPS2021_49ad23d1} & ImageNet & 1.0k \\
\begin{tabular}[c]{@{}l@{}}Auto-\\ regressive\end{tabular} & Dalle & LAION & 1.0k \\
\bottomrule
\end{tabular}
}
\end{table}

To expand the evaluation to more challenging scenarios, we further included the Synthbuster \cite{10334046} dataset. Synthbuster contains synthetic images generated from 9 different models (1000 per model): DALL$\cdot$E2, DALL$\cdot$E3, Adobe Firefly, Midjourney v5, Stable Diffusion 1.3, Stable Diffusion 1.4, Stable Diffusion 2, Stable Diffusion XL and Glide. The original images are based on the raise-1k images dataset \cite{10.1145/2713168.2713194}. For each image of the raise-1k dataset, a description was generated using the Midjourney and each of these prompts was manually edited to produce results as photo-realistic as possible.

\subsection{Integrated Models}
To map the current landscape of SID methods, we conducted a systematic literature review and identified 11 recent state-of-the-art SID models (since 2020). Our compilation considered two main aspects: a) the type of input used for detection, and b) the architecture of the proposed model. Regarding input, we can categorize models into those applied directly to raw images and those that extract fingerprints from the images, ranging from patterns in the frequency domain to texture patterns, noise patterns, and image gradients from pre-trained deep networks, among others. 
When it comes to the backbone architecture, many pre-trained networks have been used, with ResNet50 pre-trained on ImageNet, and ViT L/14, pre-trained on image-text pairs with a contrastive objective (CLIP), being the most popular. Therefore, we included models based on these two architectures in our framework. Finally, another factor is whether the model is applied on the entire image or focuses on specific local regions (patches). So, we included examples of both approaches in our study. Table \ref{tab:model_categorization} presents a categorization of the integrated models according to these criteria. The current list of integrated models is the following:

\begin{enumerate}

\item \textbf{CNNDetect} \cite{Wang_2020_CVPR}: A standard ResNet50 architecture pre-trained on ImageNet and fine-tuned  to detect ProGAN generated images with appropriate selection of data augmentations to improve generalization.

\item \textbf{LGrad} \cite{Tan_2023_CVPR}: This  uses image gradients, computed using a pre-trained deep network, as features. These are then processed by a standard ResNet50 pre-trained on ImageNet, which is fine-tuned on ProGAN images.

\item \textbf{DIMD} \cite{corvi2023detection}: Similar to CNNDetect, with a modified ResNet50 that avoids down-sampling layers to preserve high frequency fingerprints generated by GANs. It is trained on images generated by ProGAN, Latent Diffusion, and StyleGAN2 \cite{Karras_2020_CVPR}, with several augmentations for better generalization.

\item \textbf{FreqDetect} \cite{pmlr-v119-frank20a}: A standard ResNet50 architecture that employs frequency representations as fingerprints for detecting GAN-generated deep fake images, trained on ProGAN.

\item \textbf{Fusing} \cite{9897820}: An attention-based approach that fuses global and local ResNet50 embeddings from the whole image and informative patches. It is trained on ProGAN images.

\item \textbf{GramNet} \cite{Liu_2020_CVPR}: A ResNet18 architecture that leverages global image texture representations based on Gram matrices. It is trained on StyleGAN \cite{Karras_2019_CVPR} images.

\item \textbf{NPR} \cite{tan2023rethinking}: A ResNet50-based architecture that uses neighboring pixel relationships to capture the structural artifacts stemming from up-sampling operations of GANs. 

\item \textbf{UnivFD} \cite{Ojha_2023_CVPR}: A linear classification layer on top of the last layer features extracted from CLIP’s ViT-L/14 image encoder. It is trained on ProGAN images. 

\item \textbf{RINE} \cite{koutlis2024leveraging}: A CLIP-based classifier that leverages the image representations extracted by intermediate Transformer blocks of CLIP’s ViT-L/14 image encoder. It is trained on ProGAN and Latent Diffusion images. 

\item \textbf{PatchCraft} \cite{zhong2024patchcraft}: A universal fingerprint is extracted, based on the inter-pixel correlation contrast between regions with high and low texture diversity within an image.

\item \textbf{DeFake} \cite{10.1145/3576915.3616588}: A hybrid approach that utilizes CLIP's image and text encoders, fine-tuned on SID with latent diffusion images as training data. 

\end{enumerate}

\begin{table}[h]
\caption{Categorizing the integrated models based on the type of input and the backbone architecture.}
\label{tab:model_categorization}
\scalebox{0.95} {
\begin{tabular}{lc|cc}
\cline{3-4}
 & \multicolumn{1}{l}{} & \multicolumn{2}{c}{Input features} \\ \cline{3-4} 
 & \multicolumn{1}{l}{} & \multicolumn{1}{c|}{Raw Images} & Fingerprints \\ \midrule
\multicolumn{1}{c|}{\begin{tabular}[c]{@{}c@{}}Backbone\\ Architecture\end{tabular}} & \begin{tabular}[c]{@{}c@{}}ResNet\\ +\\ ImageNet\end{tabular} & \multicolumn{1}{c|}{\begin{tabular}[c]{@{}c@{}}CNNDetect,\\ DIMD\\ (ResNet18)\end{tabular}} & \begin{tabular}[c]{@{}c@{}}LGrad, GramNet\\ Fusing, NPR,\\ FreqDetect\end{tabular} \\ \cline{2-4} 
\multicolumn{1}{l|}{} & \begin{tabular}[c]{@{}c@{}}ViT\\ +\\ CLIP\end{tabular} & \multicolumn{1}{c|}{\begin{tabular}[c]{@{}c@{}}UnivFD,\\ RINE,\\ DeFake\end{tabular}} & - \\ \cline{2-4} 
\multicolumn{1}{l|}{} & Other & \multicolumn{1}{c|}{-} & PatchCraft \\ \bottomrule
\end{tabular}
}
\end{table}

\subsection{Evaluation metrics}

We evaluate integrated models with established evaluation metrics on each test dataset with respect to the True Positives (TP), True Negatives (TN), False Positives (FP) and False Negatives (FN), where Positive corresponds to the fake class. As an overall performance indicator we report accuracy (ACC) :

\[
ACC = \frac{TP + TN}{TP + TN + FP + FN}
\]
By default, we use a threshold of 0.5 on the outputs of the detection models, but a study of threshold calibration is also included. We further calculate the Precision-Recall curve, based on different threshold levels. In this paper we report the Average Precision (AP) as the weighted mean of precision at each of the $n$ thresholds: 

\[
\text{AP} = \sum_n (\text{Recall}_n - \text{Recall}_{n-1}) \cdot \text{Precision}_n
\]
SIDBench also supports computing the ROC curve and the corresponding AUC-ROC metric, but we do not report it here, as we consider AP to better reflect the real-world performance of SID models. 
We also compute and report the True Positive Rate (TPR) and True Negative Rate (TNR), to understand how well fake and real images are detected respectively. 

\noindent
\[
TPR = \frac{TP}{TP + FN},   \hspace{0.3cm}
TNR = \frac{TN}{TN + FP}
\]

\section{Evaluation Results}

Tables \ref{tab:acc_gan_dm} and \ref{tab:ap_gan_dm} present the performance scores (ACC \& AP respectively) of the integrated models across the 20 datasets from \cite{Wang_2020_CVPR} and \cite{corvi2023detection}. We utilized models trained on ProGAN data from \cite{Wang_2020_CVPR}, confirming that, as with previous works, almost all models generalize well across different GANs. For instance, UnivFD \cite{Ojha_2023_CVPR} achieves an accuracy of 90\% and higher for most GANs, except for StyleGAN and StyleGAN2, where the performance is still decent. Similarly, DIMD \cite{corvi2023detection} attained an accuracy higher than 90\% across other GANs. We observe that fingerprint-based methods like LGrad, FreqDetect, Fusing, GramNet, and PatchCraft also generalize well, albeit with variations across datasets. This demonstrates that fingerprints can capture artifacts common across all GAN models. Conversely, performance on DMs — including Guided Diffusion, Latent Diffusion, and Glide — decreases for most of the models, though exceptions such as GramNet, RINE, and PatchCraft perform surprisingly well. For example, GramNet achieved an accuracy higher than 90\% across all datasets. At the same time, ResNet-based models applied on raw images such as CNNDetect and DIMD perform poorly, with accuracy slightly higher than 50\%. It is noteworthy that, on average across all datasets, RINE is the best performing model, exhibiting also lower variation. Conversely, LGrad and GramNet, which achieved the second and third best performances respectively, are negatively impacted by specific datasets in which they perform poorly (e.g., CRN and IMLE, SITD and SAN).

\begin{figure*}[]
  \centering
  \scalebox{0.95} {
  \includegraphics[width=\linewidth]{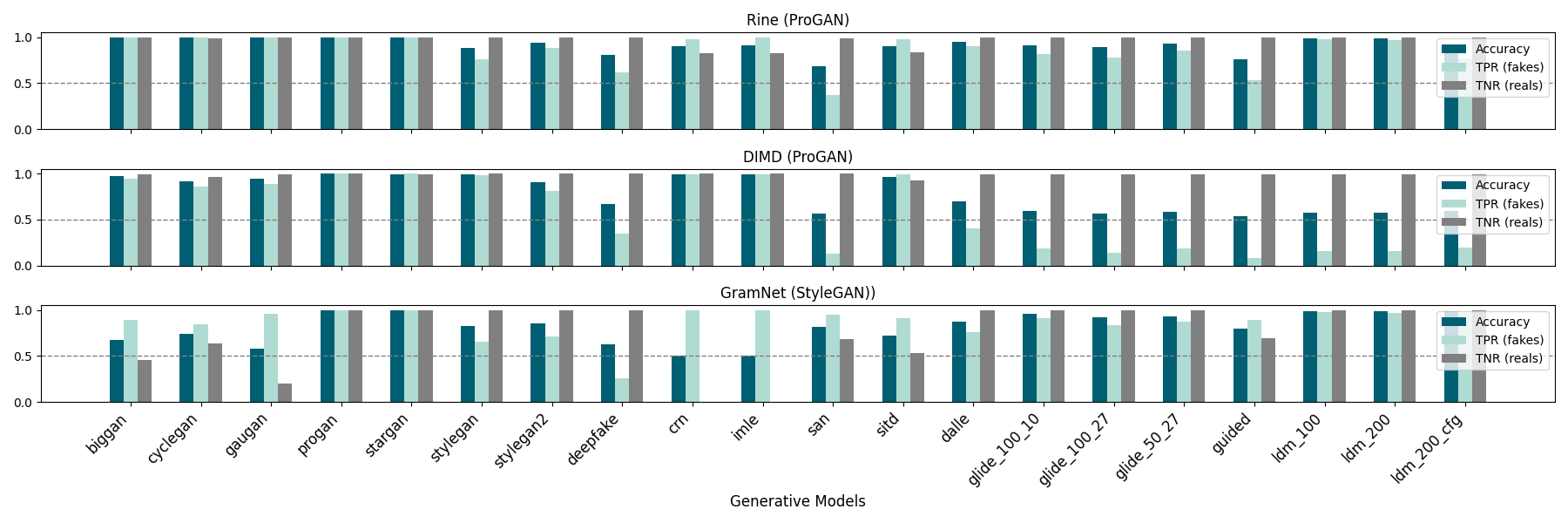}
  }
  \caption{ACC, TPR and TNR across GAN and LD datasets, for the top-3 performing models.}
    \label{fig:best_3_models_combined}
  \Description{Accuracy with best threshold compared to accuracy with default threshold (0.5)}
\end{figure*}

To understand how models perform across different datasets, we consider not only accuracy but also we calculate the TPRs (fake images) and TNRs (real images) for each model. Figure \ref{fig:best_3_models_combined} depicts these metrics for the top three performing models: RINE, DIMD, and GramNet. We observe RINE's low variation across all datasets. In datasets where it performs poorly, such as in SAN and Guided Diffusion, it is primarily the TPR that is affected, indicating the model's difficulty to detect synthetic images, while the TNR remains high, limiting false detections. Similarly, DIMD, which fails to generalize on diffusion images, exhibits a low TPR, while maintaining a consistently high TNR ($tnr \approx 0$). Conversely, GramNet demonstrates a totally different behavior in the failing cases, with the TNR being low, meaning the model classifies most real images as fakes. In certain datasets, such as CRN and IMLE, the TNR is even zero. GramNet's performance exhibits extremely high variation, from near-perfect detection to complete failure, making it inappropriate for real-world scenarios where the source and generative model of testing images is unknown. Thus, this analysis underscores the need for systematic evaluation to understand each model's performance in real-world scenarios and applications.

\begin{table*}[]
\caption{Accuracy (ACC) scores of integrated models across 20 test datasets from \cite{Wang_2020_CVPR} and \cite{Ojha_2023_CVPR}.}
\label{tab:acc_gan_dm}
\begin{adjustbox}{width=\textwidth}
\begin{tabular}{lccccccccccccccccccccc}
\toprule
 & \multicolumn{7}{c}{Generative Adversarial Networks} & \multicolumn{1}{l}{} & \multicolumn{2}{c}{Low level vision} & \multicolumn{2}{l}{Perceptual loss} & \multicolumn{1}{l}{} & \multicolumn{3}{c}{Latent Diffusion} & \multicolumn{3}{c}{Glide} & \multicolumn{1}{l}{} & \multicolumn{1}{l}{} \\ \cline{2-8} \cline{10-13} \cline{15-20}
method & \begin{tabular}[c]{@{}c@{}}Pro\\ GAN\end{tabular} & \begin{tabular}[c]{@{}c@{}}Style\\ GAN\end{tabular} & \begin{tabular}[c]{@{}c@{}}Style\\ GAN2\end{tabular} & \begin{tabular}[c]{@{}c@{}}Big\\ GAN\end{tabular} & \begin{tabular}[c]{@{}c@{}}Cycle\\ GAN\end{tabular} & \begin{tabular}[c]{@{}c@{}}Star\\ GAN\end{tabular} & \begin{tabular}[c]{@{}c@{}}Gau\\ GAN\end{tabular} & \begin{tabular}[c]{@{}c@{}}Deep\\ fake\end{tabular} & SITD & SAN & CRN & IMLE & Guided & \begin{tabular}[c]{@{}c@{}}200\\ steps\end{tabular} & \begin{tabular}[c]{@{}c@{}}200\\ CFG\end{tabular} & \begin{tabular}[c]{@{}c@{}}100\\ steps\end{tabular} & \begin{tabular}[c]{@{}c@{}}100\\ 27\end{tabular} & \begin{tabular}[c]{@{}c@{}}50\\ 27\end{tabular} & \begin{tabular}[c]{@{}c@{}}100\\ 10\end{tabular} & Dall$\cdot$E & AVG \\ 
\midrule
CNNDetect (prob 0.5) \cite{Wang_2020_CVPR} & \textbf{100.0} & 73.6 & 68.0 & 59.3 & 80.8 & 80.9 & 79.6 & 50.9 & 78.06 & 50.0 & 87.95 & 94.35 & 52.3 & 51.1 & 51.4 & 51.3 & 53.3 & 55.6 & 54.2 & 52.5 & 66.26 \\
CNNDetect (prob 0.1) \cite{Wang_2020_CVPR} & 78.70 & 86.90 & 84.60 & 52.30 & 85.65 & 92.15 & 78.70 & 53.55 & 90.28 & 50.46 & 85.95 & 85.80 & 62.00 & 53.85 & 55.20 & 55.10 & 60.30 & 62.70 & 61.00 & 56.05 & 71.54 \\
LGrad \cite{Tan_2023_CVPR} & 99.85 & 89.0 & 85.1 & 84.25 & 87.3 & 99.4 & 83.4 & 52.35 & 78.89 & 79.68 & 53.55 & 53.55 & 76.1 & 88.7 & 90.6 & 89.7 & 87.3 & 89.65 & 89.35 & 89.0 & 82.34 \\
DIMD \cite{corvi2023detection} & \textbf{100.0} & \textbf{99.1} & 90.8 & 96.8 & 91.35 & 99.35 & 94.0 & 67.15 & \textbf{96.11} & 56.62 & \textbf{99.35} & \textbf{99.35} & 53.85 & 57.55 & 59.4 & 57.6 & 56.65 & 58.7 & 58.85 & 69.65 & 78.11 \\
FreqDetect \cite{pmlr-v119-frank20a} &  99.5 & 90.8 & 72.3 & 82.2 & 79.05 & 94.4 & 81.65 & 63.85 & 66.39 & 51.14 & 59.95 & 60.05 & 57.65 & 78.95 & 76.65 & 79.25 & 52.1 & 53.3 & 49.65 & 81.5 & 71.52 \\
Fusing \cite{9897820} &  99.9 & 82.35 & 80.8 & 75.7 & 83.4 & 91.65 & 73.95 & 54.5 & 82.78 & 52.51 & 87.65 & 89.5 & 62.7 & 53.15 & 54.25 & 53.6 & 60.0 & 63.1 & 60.8 & 53.4 &70.78 \\
GramNet \cite{Liu_2020_CVPR} & \textbf{100.0} & 82.9 & 85.65 & 67.45 & 74.05 & 100.0 & 57.55 & 62.55 & 72.22 & 81.51 & 50.05 & 50.05 & 79.5 & \textbf{98.45} & \textbf{98.45} & \textbf{98.7} & \textbf{91.75} & \textbf{93.4} & \textbf{95.55} & 87.75 & 81.38 \\
NPR \cite{tan2023rethinking} & 50.0 & 50.0 & 49.95 & 50.0 & 49.7 & 50.0 & 50.0 & 54.75 & 83.06 & 50.0 & 50.1 & 50.05 & 50.25 & 49.95 & 49.95 & 49.95 & 51.55 & 52.35 & 51.45 & 50.0 & 52.15 \\
UnivFD \cite{Ojha_2023_CVPR} &  99.85 & 83.85 & 75.65 & 95.05 & 98.2 & 96.05 & 99.45 & 68.05 & 62.22 & 56.62 & 56.6 & 68.1 & 69.65 & 94.4 & 74.0 & 95.0 & 78.5 & 79.05 & 77.9 & 87.3 & 80.77 \\
RINE \cite{koutlis2024leveraging} & \textbf{100.0} & 88.0 & \textbf{94.05} & \textbf{99.5} & \textbf{99.3} & \textbf{99.75} & \textbf{99.6} & \textbf{80.3} & 90.56 & 68.26 & 90.45 & 91.45 & 76.1 & 98.25 & 88.2 & 98.6 & 88.75 & 92.55 & 90.7 & \textbf{95.0} & \textbf{91.47} \\
PatchCraft \cite{zhong2024patchcraft} & \textbf{100.0} & 91.85 & 89.3 & 95.25 & 69.05 & 100.0 & 71.2 & 56.15 & 88.06 & \textbf{88.58} & 50.05 & 50.05 & \textbf{80.5} & 91.0 & 90.05 & 90.9 & 78.9 & 82.4 & 85.2 & 85.5 & 81.7 \\
\bottomrule
\end{tabular}
\end{adjustbox}
\end{table*}

\begin{table*}[]
\caption{Average precision (AP) of integrated models across 20 test datasets from \cite{Wang_2020_CVPR} and \cite{Ojha_2023_CVPR}.}
\label{tab:ap_gan_dm}
\begin{adjustbox}{width=\textwidth}
\begin{tabular}{lccccccccccccccccccccc}
\toprule
 & \multicolumn{7}{c}{Generative Adversarial Networks} &  & \multicolumn{2}{c}{Low level vision} & \multicolumn{2}{l}{Perceptual loss} &  & \multicolumn{3}{c}{Latent Diffusion} & \multicolumn{3}{c}{Glide} &  &  \\ \cline{2-8} \cline{10-13} \cline{15-20}
\multicolumn{1}{l}{method} & \multicolumn{1}{c}{\begin{tabular}[c]{@{}c@{}}Pro\\ GAN\end{tabular}} & \multicolumn{1}{c}{\begin{tabular}[c]{@{}c@{}}Style\\ GAN\end{tabular}} & \multicolumn{1}{c}{\begin{tabular}[c]{@{}c@{}}Style\\ GAN2\end{tabular}} & \multicolumn{1}{c}{\begin{tabular}[c]{@{}c@{}}Big\\ GAN\end{tabular}} & \multicolumn{1}{c}{\begin{tabular}[c]{@{}c@{}}Cycle\\ GAN\end{tabular}} & \multicolumn{1}{c}{\begin{tabular}[c]{@{}c@{}}Star\\ GAN\end{tabular}} & \multicolumn{1}{c}{\begin{tabular}[c]{@{}c@{}}Gau\\ GAN\end{tabular}} & \multicolumn{1}{c}{Deepfake} & \multicolumn{1}{c}{SITD} & \multicolumn{1}{c}{SAN} & \multicolumn{1}{c}{CRN} & \multicolumn{1}{c}{IMLE} & \multicolumn{1}{c}{Guided} & \multicolumn{1}{c}{\begin{tabular}[c]{@{}c@{}}200\\ steps\end{tabular}} & \multicolumn{1}{c}{\begin{tabular}[c]{@{}c@{}}200\\ CFG\end{tabular}} & \multicolumn{1}{c}{\begin{tabular}[c]{@{}c@{}}100\\ steps\end{tabular}} & \multicolumn{1}{c}{\begin{tabular}[c]{@{}c@{}}100\\ 27\end{tabular}} & \multicolumn{1}{c}{\begin{tabular}[c]{@{}c@{}}50\\ 27\end{tabular}} & \multicolumn{1}{c}{\begin{tabular}[c]{@{}c@{}}100\\ 10\end{tabular}} & \multicolumn{1}{c}{Dall$\cdot$E} & \multicolumn{1}{c}{AVG} \\ 
\midrule
CNNDetect (prob 0.5) \cite{Wang_2020_CVPR} & \textbf{100.0} & 98.49 & 97.47 & 88.27 & 97.03 & 95.75 & 98.07 & 67.08 & 92.76 & 63.86 & 98.94 & 99.59 & 68.35 & 65.92 & 66.74 & 65.99 & 72.03 & 76.52 & 73.22 & 66.26 & 82.62 \\
CNNDetect (prob 0.1) \cite{Wang_2020_CVPR} & \textbf{100.0} & 99.62 & 98.83 & 84.93 & 93.88 & 98.18 & 89.89 & 89.56 & 97.23 & 70.45 & 98.49 & 98.68 & 77.67 & 71.16 & 73.01 & 72.53 & 80.51 & 84.62 & 82.06 & 71.3 & 86.63 \\
LGrad \cite{Tan_2023_CVPR} & \textbf{100.0 } & 97.96 & 98.48 & 90.9 & 95.04 & 100.0 & 94.84 & 62.44 & 89.89 & 84.84 & 73.01 & 82.6 & 85.27 & 96.98 & 97.66 & 97.23 & 94.36 & 96.07 & 96.59 & 97.1 & 91.56 \\
DIMD \cite{corvi2023detection} & \textbf{100.0} & \textbf{100.0} & \textbf{99.98} & 99.78 & 98.36 & 99.98 & 99.75 & 97.14 & 99.89 & 87.19 & \textbf{100.0} & \textbf{99.99} & 74.57 & 87.5 & 89.18 & 87.77 & 87.4 & 90.27 & 89.34 & 95.73 & 94.19 \\ 
FreqDetect \cite{pmlr-v119-frank20a} & \textbf{100.0} & 98.48 & 87.72 & 94.05 & 85.51 & 99.57 & 84.31 & 72.64 & 78.08 & 51.34 & 81.52 & 69.68 & 57.11 & 92.72 & 90.39 & 93.11 & 53.92 & 55.0 & 52.22 & 94.98 & 79.62  \\
Fusing \cite{9897820} & \textbf{100.0} & 99.24 & 99.07 & 88.91 & 93.48 & 99.33 & 88.4 & 70.24 & 88.52 & 77.66 & 92.4 & 96.31 & 80.1 & 75.32 & 77.08 & 74.87 & 83.16 & 87.91 & 85.03 & 72.36 & 86.47 \\ 
GramNet \cite{Liu_2020_CVPR} & \textbf{100.0} & 94.96 & 99.42 & 62.04 & 75.18 & 100.0 & 54.66 & 95.09 & 68.46 & 81.03 & 50.18 & 50.18 & 79.82 & \textbf{99.83} & \textbf{99.81} & 99.84 & \textbf{99.11} & \textbf{99.32} & \textbf{99.65} & \textbf{98.81} & 85.37 \\ 
NPR \cite{tan2023rethinking} & 49.66 & 57.14 & 62.79 & 47.87 & 51.08 & 49.88 & 62.85 & 61.6 & 93.61 & 80.67 & 87.46 & 95.17 & 66.51 & 39.5 & 42.08 & 39.18 & 84.58 & 89.66 & 86.96 & 42.57 & 64.54  \\ 
UnivFD \cite{Ojha_2023_CVPR} & \textbf{100.0} & 97.2 & 97.51 & 99.29 & 99.77 & 99.49 & 99.99 & 82.08 & 63.84 & 78.81 & 97.24 & 98.76 & 87.64 & 99.32 & 92.5 & 99.27 & 95.28 & 95.56 & 94.96 & 97.47 & 93.8 \\ 
RINE \cite{koutlis2024leveraging} & \textbf{100.0} & 99.47 & \textbf{99.98} & \textbf{99.99} & \textbf{99.99} & \textbf{100.0} & \textbf{100.0} & \textbf{97.82} & 97.2 & 94.96 & 97.47 & 99.67 & \textbf{96.42} & \textbf{99.83} & 98.28 & \textbf{99.87} & 98.76 & 99.24 & 98.85 & 99.33 & \textbf{98.86} \\ 
PatchCraft \cite{zhong2024patchcraft} & \textbf{100.0} & 98.87 & 97.61 & 99.38 & 84.33 & 100.0 & 81.76 & 75.29 & \textbf{100.0} & \textbf{98.54} & 59.97 & 59.89 & 90.63 & 98.32 & 97.62 & 98.41 & 89.21 & 92.48 & 94.41 & 94.29 & 90.55 \\ 
\bottomrule
\end{tabular}
\end{adjustbox}
\end{table*}

\begin{figure}[ht]
    \centering
    \scalebox{0.94}{
        \includegraphics[width=\linewidth]{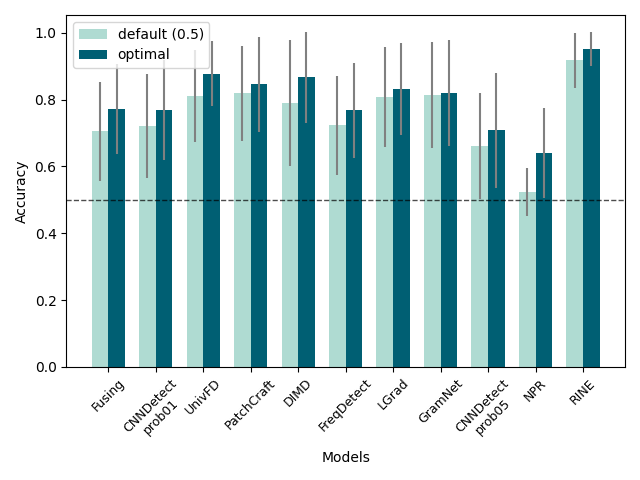}
    }
    \caption{ACC with the optimal threshold per model compared to ACC achieved with the default threshold (0.5). All models, but GramNet, have been trained on ProGAN images.}
    \label{fig:threshold_05_vs_oracle}
    \Description{ACC with optimal threshold compared to ACC with default threshold (0.5)}
\end{figure}

From the AP in Table \ref{tab:ap_gan_dm}, we can conclude that there is potential for further improvements in accuracy through calibration of the threshold used to classify images. The results in Table \ref{tab:acc_gan_dm} are based on the default threshold of 0.5 for all models and datasets. However, as observed in Table \ref{tab:ap_gan_dm}, AP is higher, in many cases exceeding 90\%. To better understand the upper limit of performance, we also calculated the accuracy using the threshold that maximizes accuracy for each model. In Figure \ref{fig:threshold_05_vs_oracle}, we present a comparison between default and oracle thresholds. It is evident that threshold calibration can lead to some improvements, but this may be a challenging task that requires additional validation data. It should be noted that the optimal threshold may vary across datasets produced by different generative models, making it challenging to find a universal threshold that performs well in real-world applications where the generative algorithm is unknown.

The datasets from \cite{Wang_2020_CVPR} and \cite{corvi2023detection}, although widely used in several works, serving as a reference for research in the field of SID, suffer from low quality and low resolution. Regarding the resolution, images are uniformly sized to 256x256 pixels. Consequently, ProGAN detectors have been trained on relatively small images. To understand the performance in real-world scenarios, we evaluated the ProGAN-trained models using Synthbuster \cite{10334046}. From Tables \ref{tab:acc_synth} and \ref{tab:ap_synth}, we observe a significant drop in performance for all models. On average, all models perform almost randomly, with accuracy around 50\%. RINE is still the best-performing model, albeit with an average accuracy of 60.26\%. It performs poorly on DALL$\cdot$E 3, Midjourney, and Glide, but achieves decent performance for Adobe Firefly, Stable Diffusion v1.3, and v1.4. Conversely, LGrad and GramNet experience an enormous performance drop. In general, fingerprint-based models (FreqDetect, Fusing) exhibit strong performance loss, indicating that artifacts existing in low-resolution images might not be present in the high-resolution, high-quality images of Synthbuster.

We can conclude that DALL$\cdot$E3 is among the most challenging datasets, as none of the methods is able to detect synthetic images generated by it, in contrast to DALL$\cdot$E2, which can be detected by the UnivFD detector. MidJourney v5 also presents a significant challenge, as none of the ProGAN-trained detectors can identify these images. For Stable Diffusion v1.3 (SDv1.3) and v1.4 (SDv1.4), UnivFD achieves decent performance (with ACC of 72.05\% and 71.6\%, respectively), while for SDv2 and SD-XL, the PatchCraft detector achieves similar performance. Adobe Firefly proves to be less challenging for RINE and UnivFD, both of which can achieve ACC values near 80\% (79.35\% and 78.15\% respectively). 

\begin{table*}[]
\caption{Accuracy (ACC) scores of integrated models for the Synthbuster \cite{10334046} dataset}
\label{tab:acc_synth}
\scalebox{0.85}{
\begin{tabular}{lllllllllll}
\toprule
 &  & \multicolumn{2}{c}{DALL$\cdot$E} & \multicolumn{4}{c}{Stable Diffusion} &  &  &  \\ \cline{3-8}
Method & Glide & \multicolumn{1}{c}{v2} & \multicolumn{1}{c}{v3} & \multicolumn{1}{c}{SDv1.3} & \multicolumn{1}{c}{SDv1.4} & \multicolumn{1}{c}{SDv2} & \multicolumn{1}{c}{SD-XL} & \multicolumn{1}{c}{\begin{tabular}[c]{@{}c@{}}Mid\\ journey\\ v5\end{tabular}} & \begin{tabular}[c]{@{}l@{}}Adobe\\ Firefly\end{tabular} & AVG \\ \hline
\midrule
CNNDetect (prob 0.5) \cite{Wang_2020_CVPR} &  48.15 & 50.5 & \textbf{47.4} & 47.75 & 47.8 & 49.35 & 49.4 & 47.9 & 53.7 & 49.11 \\
CNNDetect (prob 0.1) \cite{Wang_2020_CVPR} &  49.75 & 55.55 & 45.15 & 49.8 & 49.9 & 49.75 & 54.2 & 48.35 & 53.0 & 50.61 \\
LGrad \cite{Tan_2023_CVPR} &  \textbf{56.55} & 38.45 & 24.85 & 65.65 & 66.6 & 37.35 & 40.85 & 37.4 & 31.45 & 44.35 \\
DIMD \cite{corvi2023detection}  &  49.25 & 50.8 & 45.8 & 49.4 & 50.05 & 50.0 & 47.9 & 47.1 & 61.7 & 50.22 \\
FreqDetect \cite{pmlr-v119-frank20a} &  34.3 & 30.2 & 34.3 & 67.85 & 68.0 & 33.7 & 49.3 & 31.4 & 64.6 & 45.96 \\
Fusing \cite{9897820} &  52.7 & 54.2 & 46.75 & 50.75 & 50.85 & 49.9 & 49.6 & 49.3 & 52.5 & 50.73 \\
GramNet \cite{Liu_2020_CVPR} &  50.15 & 57.8 & 14.4 & 56.15 & 56.55 & 35.95 & 49.65 & 35.65 & 12.25 & 40.95 \\  
NPR \cite{tan2023rethinking} &  45.4 & 45.45 & 45.2 & 45.2 & 45.2 & 45.25 & 45.3 & 46.15 & 49.65 & 45.87 \\
UnivFDn \cite{Ojha_2023_CVPR} &  40.7 & \textbf{70.95} & 34.8 & 57.85 & 57.35 & 63.1 & 55.45 & 39.9 & 78.15 & 55.36 \\
RINE  \cite{koutlis2024leveraging} &  54.65 & 70.5 & 33.5 & \textbf{72.05} & \textbf{71.6} & 57.95 & 59.9 & 42.85 & \textbf{79.35} & \textbf{60.26} \\
PatchCraft \cite{zhong2024patchcraft} &  52.65 & 53.05 & 24.1 & 67.1 & 67.35 & \textbf{68.25} & \textbf{71.35} & \textbf{53.65} & 55.2 & 56.97 \\
\bottomrule
\end{tabular}
}
\end{table*}

\begin{table*}[]
\caption{Average Precision (AP) of integrated models for the Synthbuster \cite{10334046} dataset}
\label{tab:ap_synth}
\scalebox{0.85}{
\begin{tabular}{lllllllllll}
\toprule
 &  & \multicolumn{2}{c}{Dall$\cdot$E} & \multicolumn{4}{c}{Stable Diffusion} &  &  &  \\ \cline{3-8}
Method & Glide & \multicolumn{1}{c}{v2} & \multicolumn{1}{c}{v3} & \multicolumn{1}{c}{SDv1.3} & \multicolumn{1}{c}{SDv1.4} & \multicolumn{1}{c}{SDv2} & \multicolumn{1}{c}{SD-XL} & \multicolumn{1}{c}{\begin{tabular}[c]{@{}c@{}}Mid\\ journey\\ v5\end{tabular}} & \begin{tabular}[c]{@{}l@{}}Adobe\\ Firefly\end{tabular} & AVG \\ \hline
\midrule
CNNDetect (prob 0.5)\cite{Wang_2020_CVPR} & 38.32 & 51.16 & 32.74 & 38.75 & 39.26 & 46.95 & 46.16 & 36.45 & 60.77 & 43.39 \\
CNNDetect (prob 0.1)\cite{Wang_2020_CVPR} & 48.02 & 60.72 & 32.92 & 47.67 & 49.14 & 47.59 & 56.83 & 42.26 & 57.65 & 49.2 \\
LGrad \cite{Tan_2023_CVPR} & \textbf{69.23} & 41.51 & 31.89 & 63.29 & 65.36 & 39.11 & 41.67 & 39.49 & 34.71 & 47.36 \\
DIMD \cite{corvi2023detection} & 49.42 & 60.13 & 31.42 & 52.03 & 52.98 & 49.49 & 43.24 & 40.35 & 76.61 & 50.63 \\
FreqDetect \cite{pmlr-v119-frank20a} &  36.79 & 36.36 & \textbf{37.83} & 82.52 & \textbf{83.0} & 36.88 & 44.13 & 35.45 & 52.3 & 49.47 \\
Fusing \cite{9897820} & 52.8 & 58.99 & 32.73 & 49.02 & 49.32 & 49.26 & 45.7 & 43.73 & 59.0 & 48.95 \\
GramNet \cite{Liu_2020_CVPR} & 50.28 & 57.08 & 31.62 & 55.98 & 56.5 & 40.45 & 48.72 & 39.66 & 31.32 & 45.73 \\  
NPR \cite{tan2023rethinking} & 40.45 & 39.95 & 32.96 & 31.68 & 31.68 & 33.19 & 37.51 & 35.72 & 48.36 & 36.83 \\
UnivFDn \cite{Ojha_2023_CVPR} & 56.6 & \textbf{79.85} & 30.73 & \textbf{83.08} & 82.92 & 65.44 & \textbf{68.31} & 40.86 & \textbf{92.91} & \textbf{66.75} \\
RINE \cite{koutlis2024leveraging} & 54.65 & 70.5 & 33.5 & 72.05 & 71.6 & 57.95 & 59.9 & 42.85 & 79.35 & 60.26 \\
PatchCraft \cite{zhong2024patchcraft} & 48.82 & 45.79 & 32.8 & 79.21 & 81.35 & \textbf{79.33} & 64.27 & \textbf{51.06} & 44.27 & 58.54 \\
\bottomrule
\end{tabular}
}
\end{table*}

To understand the effect of the training dataset, we included in our experiments three models trained on datasets other than ProGAN. Specifically, we used DeFake \cite{10.1145/3576915.3616588}, trained on Latent Diffusion images, an instance of RINE trained on Latent Diffusion images from \cite{Ojha_2023_CVPR}, and two variants of DIMD trained on StyleGAN2 and Latent Diffusion images, respectively. In Table \ref{tab:acc_ap_other}, we present ACC \& AP values on all datasets (GAN-based from \cite{Wang_2020_CVPR}, diffusion from \cite{Ojha_2023_CVPR}, and Synthbuster \cite{10334046}). For DeFake, we were not able to replicate results similar to those reported in the paper. In most cases, performance is almost random with ACC and AP values around 50\%. However, in Synthbuster, known for its challenging nature, DeFake appears to outperform other diffusion datasets, such as those in \cite{Ojha_2023_CVPR}. Particularly in DALL$\cdot$E3, DeFake achieves an accuracy of 78.5\%, the highest among all detection methods. Moreover, an AP of 92.4\% suggests that further calibration of the classification threshold could yield even better results. Such high performance can be attributed to its focus on text-to-image DMs, and to its hybrid nature, incorporating both visual information and text captions. The high-resolution images in Synthbuster likely provide more informative captions than the lower-resolution ones. RINE, trained on Latent Diffusion, shows improvements in detecting DM images, while maintaining its good performance on GAN-based images. For example, ACC in GANs is 87.2\% on average, while in \cite{Ojha_2023_CVPR} diffusion datasets, it is 92.86\%. In Synthbuster, RINE achieved a very high 82.84\% on average, which ranks among the best scores in this challenging dataset. DIMD trained on StyleGAN2 performs better in GAN images compared to the original version trained on ProGAN. However, it fails to generalize to DM images. On the contrary, DIMD trained on Latent Diffusion exhibits better performance on images generated by other LDMs but not in Glide and Guided and shows a large performance drop in GANs.

It is noteworthy that models trained on DM images achieve improved results on some challenging subsets of Synthbuster. For instance, in Midjourney v5, which shows to be challenging according to Table \ref{tab:acc_synth}, both RINE and DIMD can achieve high accuracy when trained on LD images, with DIMD exhibiting almost perfect detection (accuracy of 98.45\%). This suggests that Midjourney incorporates a diffusion component with unique characteristics that are not detected by GAN detectors but can be identified when LD is included in the training set. Similarly, all Stable Diffusion images can be detected with high accuracy by both models.

\begin{table}[]
\caption{ACC and AP for detection models trained on Latent Diffusion (LD) and StyleGAN2 training data from \cite{corvi2023detection}.}
\label{tab:acc_ap_other}
\scalebox{0.75}{
\begin{tabular}{lclclclcl}
\toprule
& \multicolumn{2}{c}{\textbf{DeFake} \cite{10.1145/3576915.3616588}} & \multicolumn{2}{c}{\textbf{RINE} \cite{koutlis2024leveraging}} & \multicolumn{4}{c}{\textbf{DIMD} \cite{corvi2023detection}} \\ \cline{2-9} 
\begin{tabular}[c]{@{}l@{}}Generative\\ Model\end{tabular} & 
\multicolumn{2}{c}{LD} & 
\multicolumn{2}{c}{LD} &
\multicolumn{2}{c}{StyleGAN2} & \multicolumn{2}{c}{LD} \\

\midrule
\midrule
ProGAN  &  34.6 & 38.4 & 99.6 & 100.0 & 99.85 & 100.0 & 50.8 & 93.03 \\
StyleGAN  &  39.5 & 40.02 & 86.85 & 98.49 & 100.0 & 100.0 & 60.75 & 74.29 \\
StyleGAN2  &  45.1 & 46.31 & 79.55 & 85.21 & 100.0 & 100.0 & 51.5 & 73.11 \\
BigGAN  &  45.15 & 45.07 & 88.35 & 99.76 & 96.75 & 99.87 & 52.25 & 77.14 \\
CycleGAN  &  49.3 & 49.36 & 94.4 & 99.89 & 92.45 & 99.16 & 46.9 & 48.95 \\
StarGAN  &  35.5 & 37.55 & 63.6 & 99.97 & 100.0 & 100.0 & 43.45 & 45.44 \\
GauGAN  &  42.0 & 37.1 & 98.05 & 99.99 & 89.05 & 99.66 & 50.8 & 89.68 \\
\midrule
GANs Avg.  &  41.59 & 41.97 & 87.2 & 97.62 & 96.87 & 99.81 & 50.92 & 71.66 \\
\midrule
\midrule
Deepfake  &  49.25 & 49.56 & 80.5 & 93.17 & 54.65 & 97.3 & 70.05 & 87.96 \\
SITD  &  70.56 & 88.42 & 61.11 & 96.94 & 97.22 & 100.0 & 83.06 & 95.54 \\
SAN  &  66.67 & 80.74 & 67.58 & 83.09 & 50.91 & 66.09 & 81.51 & 98.29 \\
CRN  &  52.85 & 54.83 & 82.5 & 89.3 & 93.5 & 99.9 & 50.05 & 96.67 \\
IMLE  &  40.5 & 38.27 & 92.85 & 99.83 & 96.5 & 99.94 & 49.95 & 97.24  \\
\midrule
\midrule
Guided  &  47.1 & 47.18 & 93.2 & 97.96 & 58.05 & 89.21 & 51.65 & 71.39 \\
LD (200 steps)  &  41.4 & 42.88 & 95.9 & 99.99 & 68.4 & 95.39 & 99.3 & 100.0 \\
LD (200 cfg)  &  58.5 & 63.04 & 95.9 & 99.99 & 65.3 & 92.67 & 99.3 & 100.0 \\
LD (100 steps)  &  41.2 & 42.04 & 95.9 & 100.0 & 70.2 & 96.18 & 99.3 & 100.0 \\
Glide (100, 27)  &  58.55 & 63.69 & 87.3 & 95.29 & 53.9 & 74.62 & 57.8 & 83.29 \\
Glide (50, 27)  &  57.55 & 63.29 & 89.55 & 96.25 & 54.6 & 79.69 & 59.55 & 86.19 \\
Glide (100, 100)  &  59.95 & 65.27 & 89.25 & 96.22 & 54.5 & 75.08 & 62.6 & 89.6 \\
Dall$\cdot$E  &  47.0 & 47.19 & 95.9 & 99.99 & 62.95 & 91.94 & 88.55 & 98.27 \\
\midrule
Diffusion Avg.  &  51.41 & 54.32 & 92.86 & 98.21 & 60.99 & 86.85 & 77.26 & 91.09 \\
\midrule
\midrule
Glide  &  73.9 & 85.66 & 87.65 & 93.77 & 48.7 & 47.19 & 52.45 & 62.67 \\
Dall$\cdot$E2  &  41.6 & 41.59 & 81.2 & 90.2 & 48.25 & 52.66 & 49.6 & 41.7 \\
Dall$\cdot$E3  &  78.5 & 92.4 & 46.8 & 32.66 & 47.3 & 33.08 & 49.4 & 33.07 \\
SD v1.3  &  56.55 & 59.95 & 95.7 & 99.87 & 72.45 & 88.14 & 99.4 & 100.0 \\
SD v1.4  &  55.95 & 59.28 & 95.6 & 99.82 & 73.05 & 88.77 & 99.4 & 100.0 \\
SD v2  &  57.9 & 63.1 & 88.75 & 95.58 & 65.25 & 81.58 & 97.95 & 99.68 \\
SDXL  &  49.55 & 51.57 & 94.6 & 99.22 & 51.65 & 59.42 & 96.6 & 99.67 \\
Midjourney v5  &  58.9 & 63.81 & 89.2 & 95.58 & 51.35 & 59.2 & 98.45 & 99.88 \\
Adobe Firefly  &  41.05 & 40.3 & 66.1 & 74.91 & 70.45 & 88.51 & 57.6 & 86.13 \\
\midrule
Synthbuster Avg.  &  57.1 & 61.96 & 82.84 & 86.85 & 58.72 & 66.51 & 77.87 & 80.31\\
\midrule
\midrule
Total Average  &  51.59 & 55.1 & 85.64 & 93.55 & 71.97 & 84.66 & 69.31 & 83.75 \\
\bottomrule
\end{tabular}}
\end{table}

\subsection{Exploring the Influence of Image Resolution on Model Performance}

An important characteristic that distinguishes Synthbuster from the datasets of \cite{Wang_2020_CVPR} and \cite{Ojha_2023_CVPR} is the high resolution of its images. Typically, the datasets from \cite{Wang_2020_CVPR} and \cite{Ojha_2023_CVPR} consist of images with the same low resolution (256x256 pixels), while Synthbuster includes much larger images, the size of which also depends on the generative model or tool used. For example, Midjourney images are usually around 1300x800 pixels, while Adobe Firefly images exceed 2000x1700 pixels. This difference raises two important issues. First, the majority of SID models have been trained on low-resolution images (e.g., ProGAN 256x256 images), meaning that they have not encountered large images with complex visual content during the training. Second, some image encoders, such as CLIP’s ViT-L/14 used in UnivFD, RINE, and DeFake, require images to be of a specific size (in this case, 224 pixels). Therefore, to handle Synthbuster images with pretrained models, there are two options: a) either crop the images to a predefined size (center crop or random) or b) resize them. In the case of cropping, it is obvious that additional information outside the cropped area is lost, while resizing introduces a distortion that may affect performance. For example, resizing an image affects its frequency components, which could impact models that rely on frequency artifacts, such as FreqDetect.

\begin{figure}[]
    \centering
    \includegraphics[width=\linewidth]{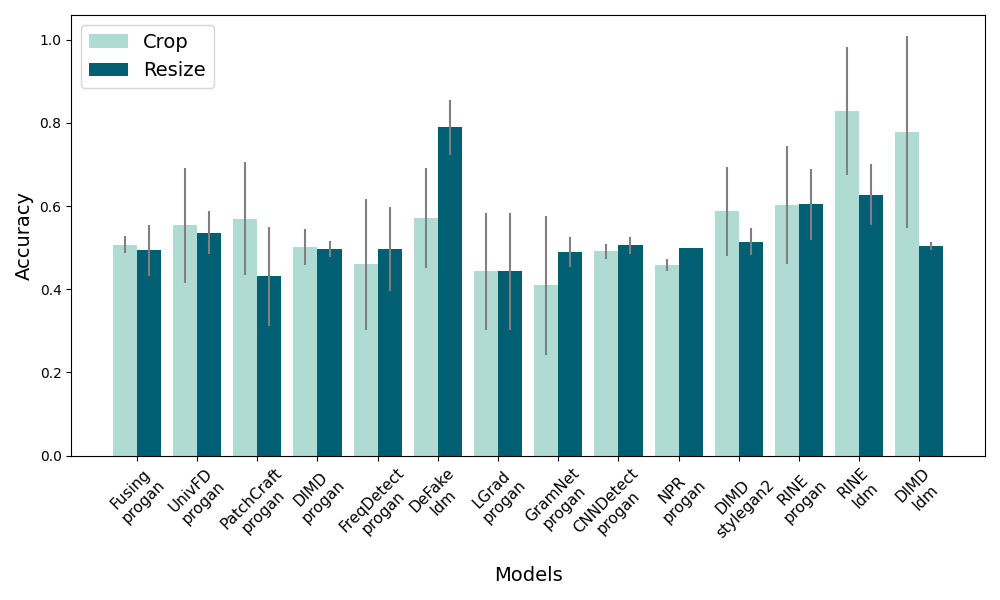}
    \caption{Comparing accuracy across cropped and resized images in the Synthbuster \cite{10334046} dataset}
    \Description[Synthbuster accuracy]{Comparing accuracy across cropped and resized images in the Synthbuster \cite{10334046} dataset}
    \label{fig:resize_impact}
\end{figure}

The experiments conducted so far have utilized cropped versions of the Synthbuster images. To evaluate whether cropping or resizing affects performance and to quantify this impact, we reran the evaluation on the Synthbuster using the resized version. For CLIP-based models, we resized images to 224 pixels, while for all other cases, we resized them to 256 pixels. Figure \ref{fig:resize_impact} presents the average accuracy for all models across the 9 Synthbuster datasets, alongside the standard deviation. It is observed that, in most cases, the average performance remains quite similar. However, the standard deviation indicates that resizing versus cropping affects the performance of each model across the Synthbuster datasets differently. For instance, the average accuracy of UnivFD or ProGAN-trained RINE is similar for both cropped and resized images, but the variation is significantly higher in the case of cropped ones. Conversely, DIMD and RINE, when trained on Latent Diffusion, exhibit much higher performance with cropped images, while their accuracy significantly drops with resized images. Considering that the former is based on ResNet50 pretrained on ImageNet and the latter on ViT-L/14 pretrained on CLIP, we can infer that it is not the encoder that is affected by cropping or resizing but the classification layers that follow it. On the other hand, in DeFake, resizing improves the average performance significantly. As DeFake, considers both images and text captions generated by the visual part resizing largely retains the available information, in contrast to cropping.

\begin{figure}[]
    \centering
    \scalebox{0.95}{
    \includegraphics[width=\linewidth]{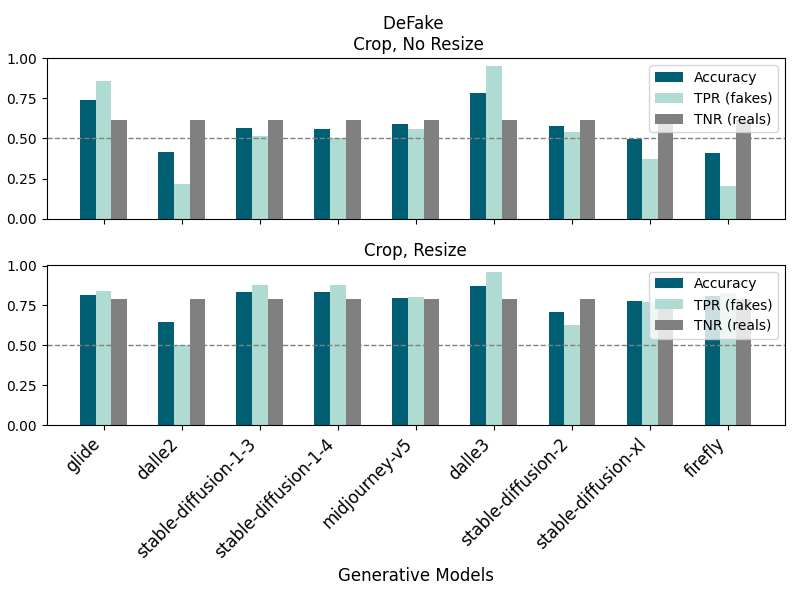}
    }
    \caption{ACC, TPR, and TNR of the DeFake model with cropped and resized images in the Synthbuster dataset \cite{10334046}}
     \Description[DeFake ACC in Synthbuster]{ACC, TPR, and TNR of the DeFake model with cropped and resized images in the Synthbuster dataset \cite{10334046}}
    \label{fig:defake_resize}
\end{figure}

To further investigate DeFake, Figure \ref{fig:defake_resize} showcases its performance across each of the 9 Synthbuster datasets, including TPR and TNR. Upon resizing, both rates improve for most datasets, with significant improvements observed in SDv1.3, and SDv1.4, SDXL, and Firefly. Compared to other CLIP models, such as UnivFD and RINE, the performance variation is considerably reduced.

\subsection{Influence of Image Transformations}

\begin{figure*}[t!]
    \centering
    \scalebox{0.95}{
        \includegraphics[width=\linewidth]{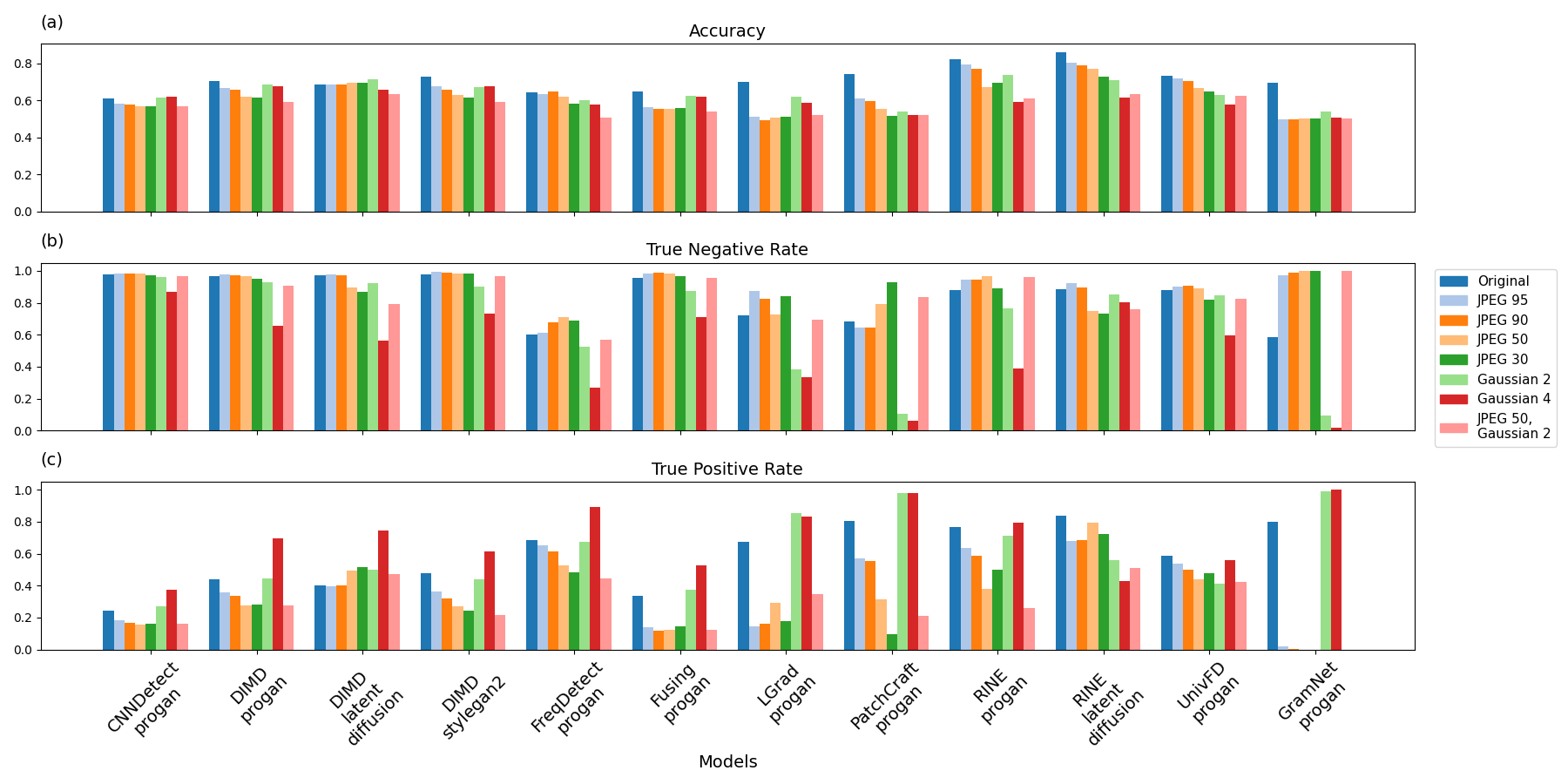}
    }
    \caption{ACC, TPR and TNR across detection models for different image transformations. Metrics have been averaged across all evaluation datasets.}
    \Description[]{}{}
    \label{fig:transformations_acc}
\end{figure*}

It is known from previous SID works that image transformations, such as the addition of Gaussian blurring, JPEG re-compression, resizing, etc., negatively affect the performance of detection models. To counteract this, a common practice is to perform strong data augmentation during training, incorporating such transformations. In this work, we tested the performance of the integrated models under two types of transformations: Gaussian blurring with varying degrees of $\sigma$ (2 and 4) and JPEG re-compression with varying image quality (95, 90, 50 and 30). Figure \ref{fig:transformations_acc} presents the ACC, TPR and TNR, when different transformations are applied. We can observe in the first subplot that the inclusion of any transformation results in worse performance in terms of ACC compared to the original images with a varying degree of impact. For some methods, such as RINE, JPEG re-compression has less impact compared to Gaussian blurring, while for others, such as DIMD (trained on ProGAN or StyleGAN), Gaussian blur has a negligible effect. 

However, accuracy alone cannot provide the full picture. For example, with GramNet, we observe that all transformations have more or less the same degree of negative impact. In contrast, for models like CNNDetect trained on ProGAN, blurring marginally improves accuracy, as it does for DIMD trained on latent diffusion. However, the TPR and TNR paint a different picture. We can observe that transformations significantly affect these metrics, effectively neutralizing each other, resulting to moderate changes in accuracy. For instance, in the CNNDetect example, the slight increase in accuracy can be attributed to the increase in the TPR (i.e., the detection of fakes), with the downside of a decrease in TNR. In other words, blurring results in more images being detected as fakes, including some real ones. This behaviour is similar for most models. TPR increase at the expense of TNR, resulting in the models detecting blurred images as fakes. This is evident in the GramNet and PatchCraft cases, where blurring results in almost all images being classified as fakes. It is intriguing that the observed behavior disappears when blurring is combined with JPEG re-compression, suggesting that this phenomenon needs more in-depth investigation. With JPEG re-compression, behavior is more inconsistent across models, as variations in the TPR and TNR differ. Generally, the TNR is less impacted than the TPR, except for GramNet. For this model, the TNR increases close to 1, while the TPR drops to 0, indicating that JPEG compression leads to GramNet's complete inability to detect any fake images.

\section{Conclusions}
The SID task presents a considerable challenge, primarily due to the increasing number of generative models and the plethora of detection models developed to counter them. No single detector emerges as a clear winner across all datasets, underscoring the complexity of the field. Given the number of approaches, finding a one-size-fits-all model that works well for all kinds of synthetic images is still an open problem. This variation in performance underscores the importance of a more systematic assessment of detection models to evaluate their effectiveness in real-world applications.
Our proposed SIDBench represents a step towards such a systematic evaluation of SID models. Recognizing that all detection models exhibit sensitivity to image transformations, such as blurring, resizing, cropping, and JPEG compression, our framework introduces comprehensive evaluation metrics to aid in the deeper understanding of each model's performance. 
In the future, SIDBench will be extended to include a wider range of datasets, particularly those reflecting real-world cases, as well as new detection models. 
Additionally, we plan to incorporate image encodings, such as .webp, and to explore new evaluation dimensions, including robustness to adversarial attacks and fairness. Finally, we will focus on combining various detection methods, e.g. through ensembles, to leverage their complementarity and enhance the overall performance. 

\begin{acks}
This work is partially funded by the Horizon Europe projects \href{https://www.veraai.eu/home}{vera.ai} under grant agreement no. 101070093 and \href{https://aicode-project.eu/}{AI-CODE} under grant agreement no. 101135437. Is is also partially supported by a research partnership project with \href{https://www.logically.ai/}{logically.ai}, who provides contributions in terms of real-world test data and subject matter expertise.
\end{acks}


\clearpage
\balance
\bibliographystyle{ACM-Reference-Format}
\bibliography{refs}

\end{document}